\documentclass{IEEEtran}
\usepackage{hyperref}
\usepackage{cite}
\usepackage{graphicx} 
\usepackage{epsfig}
\usepackage{amsmath}

\begin{document}
\title{
Macro-optimization of email recommendation response rates harnessing individual activity levels and group affinity trends
	}

\author{Mohammed Korayem, Khalifeh Aljadda, and Trey Grainger \\  
	CareerBuilder, GA. \\ 
	 \texttt{mohammed.korayem,khalifeh.aljadda,trey.grainger@careerbuilder.com} \\
}  

%


\maketitle

\begin{abstract}

Recommendation emails are among the best ways to re-engage with customers after they have left a website. While on-site recommendation systems focus on finding the most relevant items for a user at the moment (right item), email recommendations add two critical additional dimensions: who to send recommendations to (right person) and when to send them (right time). It is critical that a recommendation email system not send too many emails to too many users in too short of a time-window, as users may unsubscribe from future emails or become desensitized and ignore future emails if they receive too many. Also, email service providers may mark such emails as spam if too many of their users are contacted in a short time-window. Optimizing email recommendation systems such that they can yield a maximum response rate for a minimum number of email sends is thus critical for the long-term performance of such a system. In this paper, we present a novel recommendation email system that not only generates recommendations, but which also leverages a combination of individual user activity data, as well as the behavior of the group to which they belong, in order to determine each user's likelihood to respond to any given set of recommendations within a given time period. In doing this, we have effectively created a meta-recommendation system which recommends sets of recommendations in order to optimize the aggregate response rate of the entire system. The proposed technique has been applied successfully within CareerBuilder's job recommendation email system to generate a 50\% increase in total conversions while also decreasing sent emails by 72\%. 
\end{abstract}

\section{Introduction}
Recommender systems are widely deployed across many industries for diverse use cases such as e-commerce, advertising, media distribution, and job boards. Recommender systems
automate the process of discovering the interests of a user and subsequently suggesting what should be relevant to his/her needs \cite{konstan2004introduction,sarwar2001item}.

Many companies depend on recommender systems to help drive their
revenue, like Netflix%
\footnote{http://www.netflix.com%
}, Amazon%
\footnote{http://www.amazon.com%
}, CareerBuilder%
\footnote{http://www.careerbuilder.com%
}, etc. For example, Netflix, a movie rental and video streaming web
site, offered a prize (known as the Netflix prize) of 1 million dollars
in 2006 for any recommendation algorithm that could beat their recommender system,
named Cinematch\cite{bennett2007netflix}. Netflix, like many other
websites, depends heavily on recommendations in order to keep their customers interested in their service. Recommendations can take place while a user is browsing a website, or even asynchronously while the user is not actively online. In the former case, recommendations are focused on selecting similar items based on the other items with which the user has previously interacted. In the latter case, recommendation emails are often sent on a regular basis (i.e. nightly, weekly, or monthly) in order to recapture offline users' attention and have them return to the website to reengage. These offline recommendation emails are more complicated than real-time recommendations due to the fact that they have to deal with two additional dimensions: 1) the right users to target among all users, and 2) the right time to send the recommendations to those users. Looking to the recommendation email process across a three-dimensional space of right person (\textit{who}), right item (\textit{what}), and right time (\textit{when}) gives us a model to optimize the effectiveness of the system at generating quality recommendations that successfully re-engage offline users.

Whereas a real-time recommendation system provides online recommendations to users while they are  browsing a website or otherwise interacting with a system, e-mail recommendations must be much more carefully optimized to ensure they are only sent to users when the user will appreciate and benefit from them. There is a fundamental supply and demand problem here: once a user has left a website, they then become your supply of potential future customers to reengage, but every time you attempt to email them, you risk having them unsubscribe if they are for any reason unhappy with the email. Sending them a timely and relevant recommendation email is a good way to get them to return and reengage, but if you send too many emails to a user then you may annoy them and lose them as a future customer forever. You wouldn't, for example, want to send a recommendation to every user of your system multiple times per day, as you may quickly run out of users to send to once they all unsubscribe from your service. Instead, it is important to drive as many successful conversions as possible with as few emails as possible, such that both you and your customers maximize the impact of your interactions.

In this paper we describe a novel system to address the three dimensions of a recommendation email system in order to maximize the aggregate response rate through sending the right content to the right people at the right time. This system has been deployed in production as part of CareerBuilder's recommendation email system, significantly increasing email response rates (by 50\%) while simultaneously reducing (by 72\%) the number of sent e-mails necessary to achieve those improved response rates.
\section{Related Work}

The main task of a recommendation system is to provide users with relevant content suggestions.  It works by collecting the preferences of users for a set of items and then ranking other items for each user based on how interested the system predicts a user will be to see those other items~\cite{bobadilla2013recommender}. There are two major types of recommendation systems: Collaborative filtering~\cite{herlocker1999algorithmic} and content-based recommendations (often refered to as content-based filtering)~\cite{pazzani2007content,bobadilla2013recommender,melville2002content}. Content-based recommendation systems recommend items for a user based on the similarity of features between a user and the items being recommended. For example, a job posting may contain features such as a job title, skills, salary, and location, and a job seeker will similarly have a desired job title, list of skills, salary, and location. Because a content-based recommendation system is just performing a similarity calculation on features, a content-based recommendation system can actually match between any two sets of entities (i.e. item to item, user to item, user to user, etc.) with a shared feature space, as it relies on no past interactions with the items by users in order to make the recommendations.

Collaborative filtering~\cite{herlocker2000explaining,Adomavicius2005,NIPS2015_5938,Su2009}, on the other hand, is based on the concept that users with a shared interest in some items will also have a shared interest in other items. For example, a user who applies to a software engineering job is likely to apply to other jobs related to software engineering, whereas a user who applies to a registered nurse job is likely to apply to other jobs related to nursing. Thus if a new user applies to a registered nurse job, there is a good chance that if we look at the other people who applied to that job and recommend the other jobs those people applied to, that our new user may be interested in those other jobs, as well. Collaborative filtering can be performed using different approaches like factorization-based methods~\cite{liu2016kernelized},
graph methods~\cite{aggarwal1999horting}, genetic algorithms ~\cite{bobadilla2011improving}, and case-based reasoning~\cite{hayes2001case}. Hybrid approaches also exist, which can combine both collaborative filtering and content-based recommendations into a unified recommendation algorithm~\cite{de2010combining,burke2002hybrid,lekakos2008hybrid,yao2013recommending}.  

While the majority of published research on recommendation systems focuses on some form of content-based, collaborative filtering-based, or hybrid algorithm for matching users with the best items ("what" to match), the additional dimensions of "when" to match and even "who" to match to maximize the response rate of the overall system are far less studied. The most related prior research to ours is ~\cite{wang2013time}, which tries to find the best time to send a job recommendation to a user in order to optimize the odds of that user acting upon the recommendation. Their system is focused on the \textit{when} component of the recommendation system, while ours combines the three components of the recommendation together (\textit{who}, \textit{what}, and \textit{when}), with a particular emphasis on the \textit{who} and \textit{what} being decided relative to a more prescriptive \textit{when} dimension.

\section{Methods}

Our methodology aims to address the three dimensions of recommendation email relevancy: who to send to, what to send, and when to send. We ultimately choose one of these dimensions - when to send, as our fixed dimension from which we will pivot, choosing to calculate who to send to and what to send relative to each time window in which we choose to send a batch of recommendations. We address these three dimensions by utilizing both individual user behavior, as well as historical group behavior from other users within the same classification, in order to figure out who to send to and what to send for each time period. The individual user behavioral data predominantly dictates who to send to relative to when the recommendations will be sent, with the goal being to maximize response rate. The group behavioral data primarily determines what to send to a particular user from a list of candidate recommendation lists in order to maximize response rate. This fusion of personal and group behavioral data provides us with better understanding of what to send, when to send, and whom to target in order to maximize the aggregate response rate across the entire batch of sent recommendation emails.

\subsection{Response Likelihood}

One of the most important goals for any recommendation email system should be to achieve a high response rate. Sending a high volume of emails with a low response rate will likely lead to several problems. First, sending too many emails to end users may overwhelm or annoy them, hurting the sender's reputation and likely resulting in the user unsubscribing from future emails or possibly breaking ties completely with the sender. Second, sending emails that are not sufficiently interesting to the end user will result in reputational harm for the sender, since the sender will be perceived as having a low-quality platform that is not worth the end user's time. Third, if too many emails are sent by the system to a particular email service provider, that email service provider may determine that the large volume of emails are spam, and they may blacklist the sender such that future emails to any recipients are blocked. Fourth, sending many emails without a good response rate is waste of resources since sending recommendation emails requires servers, queues, databases, and bandwidth to store, transmit, and track all of the emails.

With the risks of losing customers, losing the right to continue contacting customers, having a sender's reputation damaged, having all future email communications blocked to all users, and wasting resources sending ineffective emails, it is clearly important that a recommendation email system optimize how it sends emails to maximize impact while minimizing emails sent.

In order to address these issues, recommendation emails should be sent only to the users who are most likely to respond. The immediate challenge, therefore, becomes how to predict those users with high response likelihood for a set of recommendations. In our system we utilize each user's recent behavioral data with the hypothesis that active users were more recently interested and therefore are more likely to respond in general. Following this logic, if a user was active in the last 24 hours his response likelihood will be higher than the someone who was active 7 days ago, while users with a last activity within 7 days are more likely to respond than others who were active 20 days ago. For Careerbuilder we tracked three kinds of behaviors to calculate recent activity levels, namely, searching for a job, applying to a job, and updating a resume.

We use the most recent of these activities to calculate what we call the \textit{Activity Score}, which is the basis for our identification of users who are, in general, most likely to respond to recommendation emails. Assume a user $u$ was active on a date $d_{a_u}$ while today's date is $d_{t}$. Also, assume that we only consider users who were active more recently than a given date $d_{o}$. We can calculate the Activity Score $AS(u)$ as follows:
\[
AS(u) = 1 - \frac{d_{t}-d_{a_u}}{d_{t}-d_{o}}
\] 

The only parameter that needs to be chosen here is the $d_{o}$. For CareerBuilder's use case, we found that 90 days before the current date is a good cut-off threshold, as users tend to be much more responsive to recommendation emails within their first 90 days, but a response becomes much less likely beyond 90 days. 


\subsection{Group Trends}

In addition to looking at a user's Activity Score to predict the general likelihood of that user responding to a recommendation email, it is also important to consider that not all recommendation emails sent to the user are equally likely to receive a response. Because we often have limited behavioral data for any particular user regarding the various data classification for which the user may be interested, we instead rely on historical group behavioral trends to learn affinities between the group of users within each classification and their likelihood of responding to a recommendation within any other classification.  

The underlying theory here is that when users are classified into categories based on their common features, they tend to share similar interest as the other users in the same category. We built our second module of the proposed system upon that hypothesis, utilizing the group behavior within each category to predict the response rate of new users within the same category when shown items from any other category. The importance of this module is that it expands the selection pool of items beyond just those that fall under the same category as the targeted users.  Without some notion of how related different categories are, it can be risky to recommend items not within the same category as the user. For example, in the recruitment domain users who are classified into the ``Java Developer''  category would probably not have a high response rate to recommended jobs from the ``Registered Nurse''  category, while they might respond very favorably to recommendation from the ``Software Engineer''  category, and reasonably well to recommendations from the ``Hadoop Developer''  category. The recommendation engine should thus be able to understand these category affinities when predicting response rates.

To understand the probabilistic model which we implemented to represent these category affinities and to predict the interest of a user based on his group's behavior, let us first understand the notations we use to describe the model.
\begin{enumerate}

	\item $u = 1, 2, ..., U$ is the index of the user.
	\item $a,b = 1, 2, ..., C$ is the index of the category.
	\item $u_a = 1, 2, ..., U_a$ is the index of the user in category $a$.
	\item $i_a = 1, 2, ..., I_a$ is the index of the item in category $a$.
	\item $t= 1, 2, ..., T$ is the index of transition ${a\rightarrow b}$, where a user $u_a$ interacts with item $i_b$
	\item $O_t = \left\{x_{t,1}, ..., x_{t,i}\right\}$ A set of all transitions of type $t$. $x_{t,i}$ is an instance $i$ of transition $t$. 
	\item $r_{a,b}$ is the trend of users in group $a$ towards items of group $b$
	\item $S_{a,b} = \left\{x_{a,b,1}, ..., x_{a,b,n}\right\}$ A set of all items of category $b$ seen by users of category $a$.
\end{enumerate}
\begin{figure}
	\centering \epsfig{file=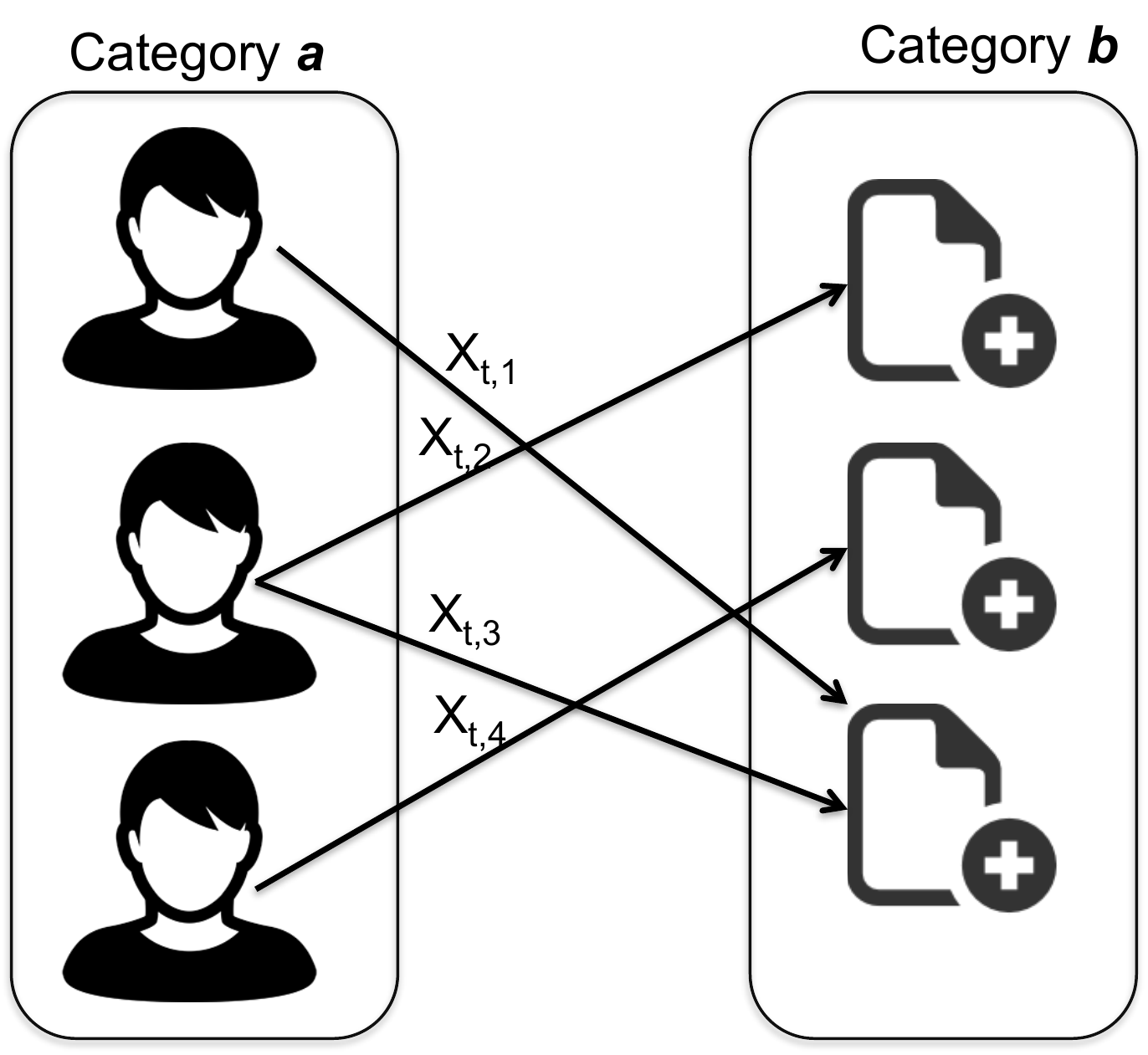,scale=0.4} \protect\caption{\label{trans} Group Trend. In this example users of category $a$ interact with items of category $b$}
	
\end{figure}

\begin{figure}
	\centering \epsfig{file=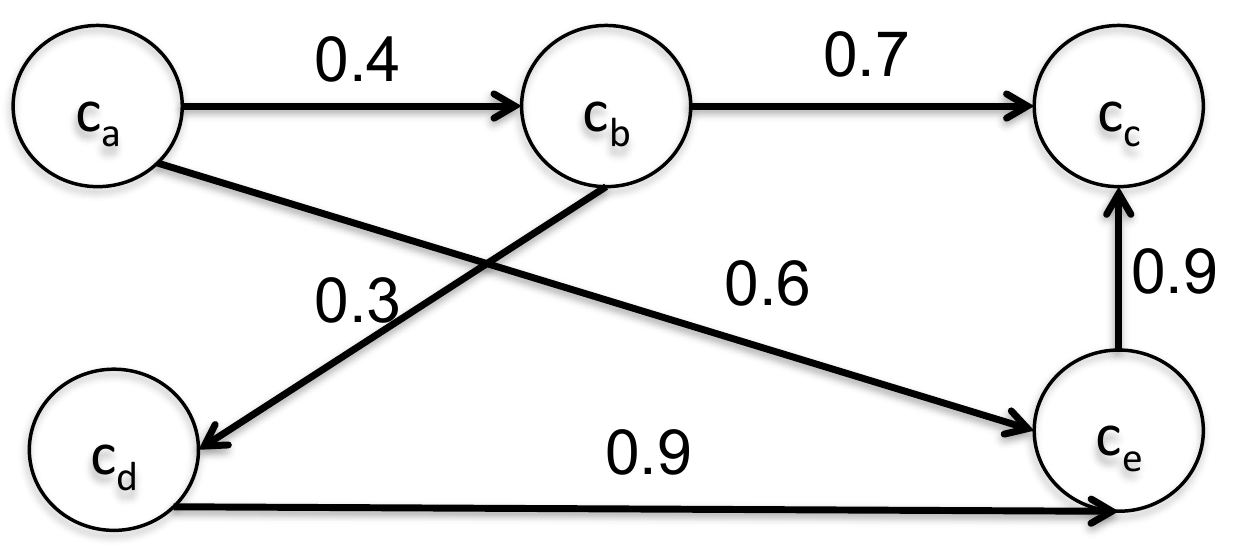,scale=0.6} \protect\caption{\label{trends} Trends Graph where we store all the possible transitions between different categories based on the calculated probability score}
\end{figure}

In figure \ref{trans} we show an example of a group trend. The example shows the transition trend of users $u_a$ towards items $i_b$ where $x_{t,1}$ is an instance of that transition and $O_t=\left\{x_{t,1}, x_{t,2}, x_{t,3}, x_{t,4}\right\}$.
We calculate the group trend as a transition probability from the users' category to another category based on the number of interactions between those users and items from the other category. 
\[
P(r_{a,b}|t,s_{a,b}) = \frac{|O_t|}{|s_{a,b}|}
\]
The probability score represents the likelihood that users from category $a$ would accept and interact with recommendations including items from category $b$. We build a transition graph modeling the probability score between different categories, and this transition graph is then utilized to select a list of recommendations corresponding with the highest probability of interest, as shown in figure \ref{trends}.
\section{Experiment and Results}

\begin{figure*}
	\centering \epsfig{file=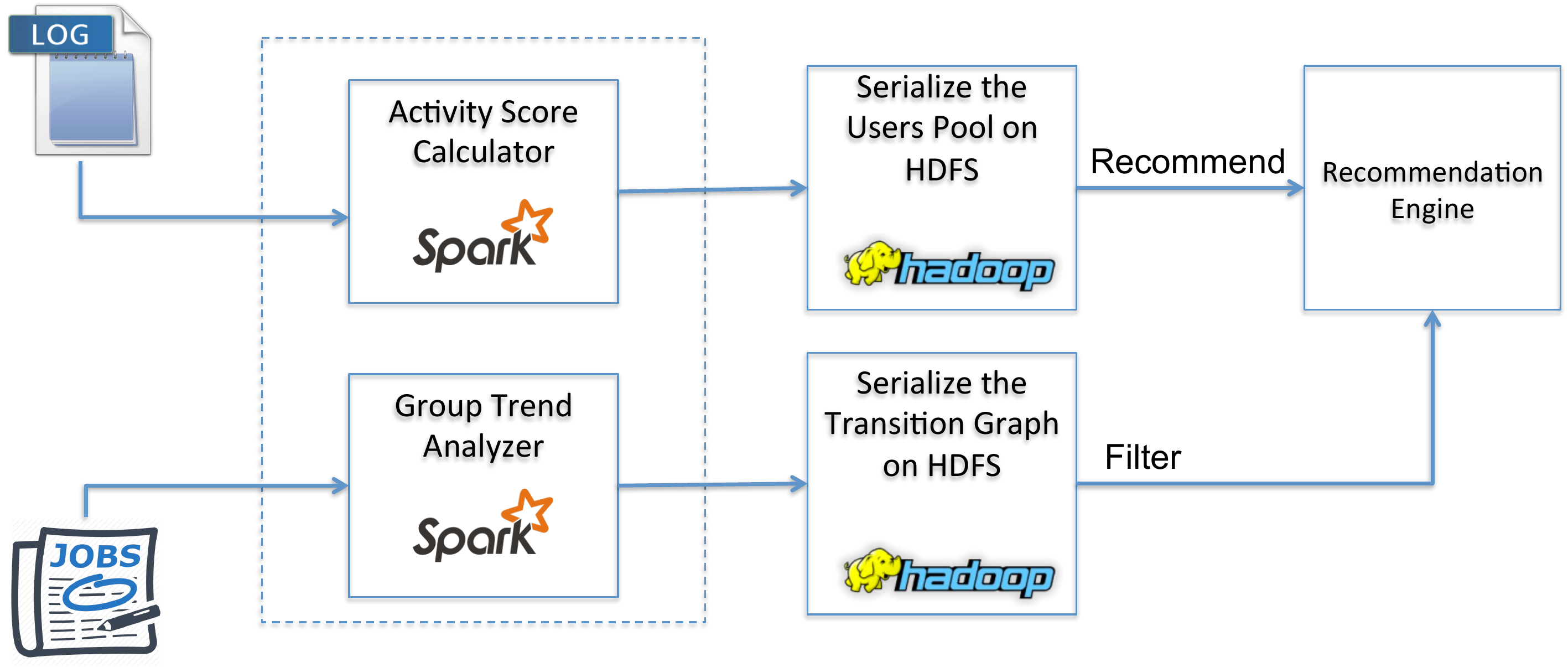,scale=0.50} \protect\caption{\label{sys} System Architecture. We leverage Apache Spark to determine each user's likelihood to respond (Activity score) to any job, as well as to analyze the group behavior of other users within each classification to determine how they typically respond to jobs within other classifications, combining these factors to determine each user's overall likelihood to respond to any recommendation made by the hybrid user-item recommendation system.}
	
\end{figure*}
\begin{table*}
	\begin{center}
		\caption{ {Results for the proposed system against the baseline (control) system. The proposed system improves Total Apps, $OSR$, $CTR$, and $AOR$ while reducing the total number of sent emails}}
		\label{results}
		\vspace{5px}
		\begin{tabular}{||c c c ||} 
			\hline
			& Baseline & Proposed System \\
			\hline
			Total Apps & 10000 & 15000\\
			\hline
			Total Sent & 500,000 & 150,000\\
			\hline
			OSR: Open to Send Ratio. Opens / Emails Sent & 29\% & 40\% \\
			\hline
			CTR: Click Through Rate. Clicks / Emails Sent & 8\% & 32\% \\
			\hline
			AOR: Apps to Open Ratio. Applications / Emails Opened  & 6\% & 25\% \\
			\hline
		\end{tabular}
	\end{center}
\end{table*}

To test the proposed system, we applied it within the recommendation email system at CareerBuilder, which is one of the largest job boards in the world. This system has millions of job postings, more than 60 million actively-searchable resumes, over one billion searchable documents, and more than a million searches per hour~\cite{aljadda2014pgmhd,aljadda2014crowdsourced,korayem2015query}. 
The recommendation engine selects jobs of interest to job seekers and then sends those jobs via recommendation emails to job seekers. Hence, these are user-item recommendations, where the users are job seekers and the items are jobs. The previous recommendation email system at CareerBuilder would restrict sent emails to the target user's category, which was not performing well because the system was overly restrictive and was unable to consider alternate, related categories that users within the initial category may also find interesting.  Our methodology (shown in figure \ref{sys}) has been applied in order to improve the quality and performance of CareerBuilder's recommendation emails, so it was important to measure how the new system is performing compared to the old one. Our measurement was based on the open to send ratio and the number of job applications created based on the recommendation emails.


We define an indicator function as 
\[ O(e_{u_i}) =
\begin{cases}
1       & \quad \text{if } e_{u_i} \text{ is opened}\\
0       & \quad \text{if } e_{u_i} \text{ is not opened}\\
\end{cases}
\]
Then we calculate the open to send ratio as:
\[
OSR = \frac{\sum\limits_{i=1}^{n}O(e_{u_i})}{n}
\]
where $OSR$ is the open to send ratio, $e_{ui}$ is the recommendation email sent to the user $u_i$, and $n$ is the total number of emails which were sent. This score represents the relevancy of the emailed job recommendations given the hypothesis that a user will not open a recommendation email if the job in that email is not of interest to that user.

While the OSR is a good initial indicator of relevancy, we should note that the user is only exposed to limited information about the job being recommended (the job title) when reviewing the subject of the email. As a result, we capture another intermediate metric called the CTR (click-through ratio). For the CTR, we first define another indicator function as 
\[ C(e_{u_i}) =
\begin{cases}
1       & \quad \text{if } e_{u_i} \text{ is clicked from a link in the email}\\
0       & \quad \text{if } e_{u_i} \text{ is not clicked from a link in the email}\\
\end{cases}
\]

Then we calculate the click-through ratio as: 
\[
CTR = \frac{\sum\limits_{i=1}^{n}C(e_{u_i})}{n}
\]

Both the OSR and the CTR provide valuable information about users' perceptions about the relevance of the recommendations they are receiving. When comparing our baseline/control algorithm versus the proposed system, these metrics also show us useful information about drop-off at each step at which the user interacts with the recommendation. Our end goal, however, to actually convert a user's interest in the job to an application for the job. To measure this, we need one additional metric: the application to open ratio (AOR). For the AOR, we defined one more indicator function:
\[ A(App_{e_{u_i}}) =
\begin{cases}
1       & \quad \text{if } App_{e_{u_i}} \text{ is a resulting job application}\\
0       & \quad \text{if } App_{e_{u_i}} \text{ is not a resulting job application}\\
\end{cases}
\]
Then we calculate the application to open ratio (AOR) as:
\[
AOR = \frac{\sum\limits_{i=1}^{n}A(App_{e_{u_i}})}{m}
\]
where $App_{e_{u_i}}$ is a job application created based on the recommendation email $e_{u_i}$, and $m$ is the total number of emails where $A(e_{u_i}) = 1$. While $OSR$ represents the open rate of the sent emails, $AOR$ represents the conversion of a recommendation e-mail into a job application, so $AOR$ is the most important factor in our case.
Table \ref{results} shows the significant improvement in $OSR$ and $AOR$ delivered by the new system over the old one.

We can essentially view the email recommendations responses as a funnel, where all users who are sent recommendations are the starting volume, less users open the email (measured by the OSR metric), even less users click on a recommendation (measured by the CTR metric), and even less users apply to the job (measured by the AOR metric). In this funnel, we note that the improvement of the proposed system over the baseline/control system compounds at each step in the funnel. For example, we note that the OSR increases from 29\% to 40\%, meaning that more users are reading the subject of the email (which lists the title of the job being recommended) and identifying it as a potentially good match based upon that limited information. Then, once a user actually views the additional information about the job in the email, his/her interest in clicking on the job (CTR) to continue engaging with it improves even further, from the baseline of 8\% CTR all the way to 32\% CTR. Finally, for each of the opened emails, we also see an improvement in actual application rate (AOR) from 6\% to 25\%, meaning that drop-off has decreased at every stage in the funnel and that users are collectively finding the recommendations from the new system more relevant and timely for their interests.

\section{Conclusion}

In this paper we presented a novel approach to improving the quality and response rate of recommendation emails. The proposed system utilizes personal behavioral data to calculate a per-user activity score to determine which users are most likely to respond at the present time based upon the most recent activity and type of activity they have exhibited. The new system additionally utilizes historical group behavior to build a transition graph which represents the probabilities that a typical user within any specific category would be likely to respond to recommendations for items from any other category. By leveraging both the group transition graph (likelihood of a typical user within a category to respond to recommendations within any particular category) and the personal activity score (likelihood of a specific user to respond to any recommendation), the proposed model is able to optimize the aggregate choice of which recommendations should be sent to which users for the given time period. The proposed model has been applied successfully within CareerBuilder's job recommendation email system to increase total conversions by 50\% while simultaneously decreasing emails sent by 72\%.

\bibliographystyle{unsrt}
\bibliography{sigproc}

\end{document}